\documentclass[10pt,twocolumn,letterpaper]{article}

\usepackage{iccv}
\usepackage{times}
\usepackage{epsfig}
\usepackage{graphicx}
\usepackage{amsmath}
\usepackage{amssymb}

\usepackage[pagebackref=true,breaklinks=true,letterpaper=true,colorlinks,bookmarks=false]{hyperref}

\usepackage{mmstyles}
\usepackage{multirow}
\usepackage[ruled]{algorithm2e}
\usepackage{floatrow}
\usepackage{appendix}

\usepackage{amsmath,amsfonts,bm}
\usepackage{booktabs}
\usepackage{multirow}
\usepackage{caption}

\def\eqref#1{equation~(\ref{#1})}

\def\1{\bm{1}}

\DeclareMathAlphabet{\mathsfit}{\encodingdefault}{\sfdefault}{m}{sl}
\SetMathAlphabet{\mathsfit}{bold}{\encodingdefault}{\sfdefault}{bx}{n}

\usepackage{float}

\newcommand\blfootnote[1]{%
  \begingroup
  \renewcommand\thefootnote{}\footnote{#1}%
  \addtocounter{footnote}{-1}%
  \endgroup
}

\iccvfinalcopy 

\def\iccvPaperID{3835} 

\ificcvfinal\fi

\begin{document}

\title{Infinite Mobility: Scalable High-Fidelity Synthesis of Articulated Objects via Procedural Generation}

\author{
        Xinyu Lian\textsuperscript{\rm1,\rm2}, 
        Zichao Yu\textsuperscript{\rm3}, 
        Ruiming Liang\textsuperscript{\rm7, \rm8}, 
        Yitong Wang\textsuperscript{\rm5}, 
        Li Ray Luo\textsuperscript{\rm1}, 
        Kaixu Chen\textsuperscript{\rm1,\rm4},\\
        Yuanzhen Zhou\textsuperscript{\rm1},
        Qihong Tang\textsuperscript{\rm6},
        Xudong Xu\textsuperscript{\rm1},
        Zhaoyang Lyu\textsuperscript{\rm1\dag},
        Bo Dai\textsuperscript{\rm9},
        Jiangmiao Pang\textsuperscript{\rm1}
        \\\\
		\textsuperscript{\rm1}Shanghai Artificial Intelligence Laboratory \textsuperscript{\rm2}South China University of Technology \\
        \textsuperscript{\rm3}University of Science and Technology of China \textsuperscript{\rm4}Tongji University\\ 
        \textsuperscript{\rm5}Fudan University\textsuperscript {\rm6}Harbin Institute of Technology, Shenzhen\\
        \textsuperscript{\rm7}Institute of Automation, Chinese Academy of Sciences \\\textsuperscript{\rm8}School of Artificial Intelligence, University of Chinese Academy of Sciences\\
        \textsuperscript{\rm9}The University of Hong Kong\\
		\tt\small\{lianxinyu,luoli,chenkaixu,zhouyuanzhen,xuxudong,lvzhaoyang,pangjiangmiao\}@pjlab.org.cn\\
        \tt\small {zichaoyu@mail.ustc.edu.cn}, \tt\small {liangruiming2024@ia.ac.cn}\\ 
        \tt\small {wangyitong23@m.fudan.edu.cn}, \tt\small {210810616@stu.hit.edu.cn}\\ 
        \tt\small {bdai@hku.hk}
	}
\ificcvfinal\thispagestyle{empty}\fi%
\twocolumn[{
\renewcommand\twocolumn[1][]{#1}
\maketitle
    \begin{center}
        \includegraphics[width=0.9\textwidth]{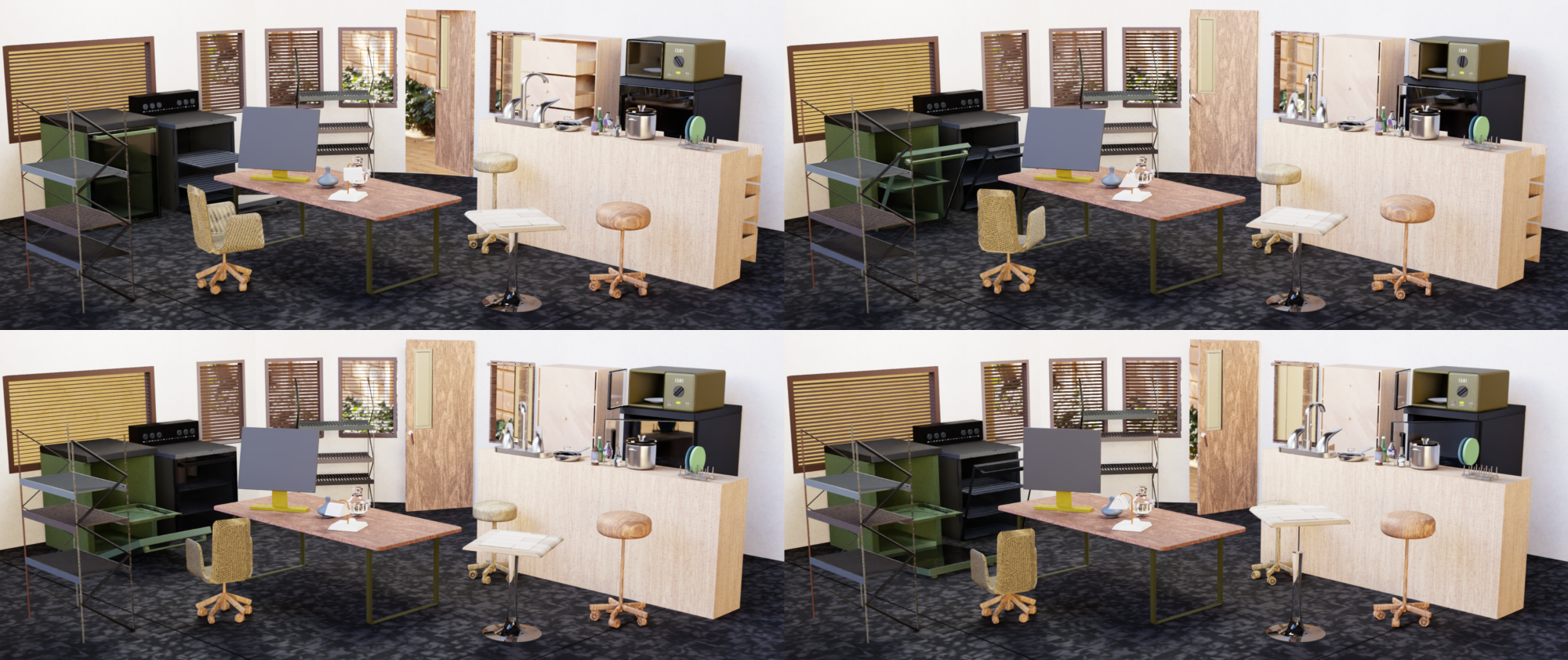}
        \captionof{figure}{
        We design probalistic programs to generate $22$ common articulated objects.
        We demonstrate the motion sequence of the generated articulated objects. 
        Ours generated articulated objects bear accurate geometry, realistic textures, and reasonable joints.
        }
    \end{center}
}]

\begin{abstract}
\vspace{-1em}
\blfootnote{
$^\dag$ Project Lead\\
\indent \indent Correspondence to: Zhaoyang Lyu, Xudong Xu
}
Large-scale articulated objects with high quality are desperately needed for multiple tasks related to embodied AI.
Most existing methods for creating articulated objects are either data-driven or simulation based, which are limited by the scale and quality of the training data or the fidelity and heavy labour of the simulation.
In this paper, we propose Infinite Mobility, a novel method for synthesizing high-fidelity articulated objects through procedural generation. 
User study and quantitative evaluation demonstrate that our method can produce results that excel current state-of-the-art methods and are comparable to human-annotated datasets in both physics property and mesh quality. 
Furthermore, we show that our synthetic data can be used as training data for generative models, enabling next-step scaling up.
Code is available at \href{https://github.com/Intern-Nexus/Infinite-Mobility}{https://github.com/Intern-Nexus/Infinite-Mobility}
\end{abstract}

\section{Introduction}
\label{sec:intro}




The scaling law has demonstrated its efficacy across diverse machine learning
domains, including computer vision, natural language processing, and
reinforcement learning. Yet, scaling the training of embodied AI-related
tasks\cite{Xiang_2020_SAPIEN,habitat19iccv,szot2021habitat} remains challenging, largely due to the scarcity of high-quality
training environments. Sim-to-real transfer\cite{8202133} offers a promising solution by
enabling training in simulation and then transferring the results to reality.
Articulated objects, such as cabinets with drawers and fridges with openable
doors, are common in the real world. They can depict objects' fine-grained
structures and the corresponding manipulation operations, making them crucial
for simulating realistic action chains in virtual environments. Although these
objects are vital for the sim-to-real paradigm, there are limited datasets
of them, and most existing datasets \cite{Xiang_2020_SAPIEN,RBO,liu2022akb48realworldarticulatedobject,kim2024parahome,NEURIPS2022_3b3a83a5} are relatively small in scale.

To create sufficiently realistic digital twins of articulated objects for sim-to-real transfer, it is essential to have both precise physical properties of joints and high-quality 3D assets.
It's natural to try to create a real-to-sim-to-real loop by reconstructing
articulated digital twins from real world objects.
But such methods\cite{RBO,
    jiang2022ditto,jiayi2023paris,liu2022akb48realworldarticulatedobject,NEURIPS2022_3b3a83a5}
need manually collected data which is labor-intensive and time-consuming.
Quality of 3D assets from these reconstructed objects are also not guaranteed.
Some try to articulate existing 3D
assets\cite{Xiang_2020_SAPIEN,wang2019shape2motion,10.1145/3355089.3356573}.
However, most of such methods rely on existing or self-made datasets for priors
of articultaion information and thus are still limited by data scale.
In recent years, generative models\cite{lei2023nap,liu2024cage} have been
adopted to generate articulated objects.
These models leverage popular diffusion
models\cite{ho2020denoisingdiffusionprobabilisticmodels} as generation
backbones to capture the training dataset distribution, but their performance remains suboptimal as most datasets contain only dozens of samples per category.
It remains a challenge to obtain large amounts of high quality articulated 3D
objects in a efficient way.

In this paper, we introduce Infinite Mobility, a procedural pipeline for
synthesizing large-scale articulated objects.
For each object, we represent its articulation as a tree structure analogous to
a URDF, where nodes correspond to links and edges represent joints.
Our approach leverages a tree-growing strategy for articulation structure generation.
Starting from a root node, we iteratively grow the tree by attaching new nodes
to existing ones.
Based on the semantic characteristics of each node, we have devised a set of
growth rules that decide whether to attach a new part and specify which
component to add if needed.
This strategy ensures that essential components are generated while
non-essential parts are incorporated gradually, facilitating the reliable and
diverse generation of articulation structures.
In contrast to above methods which rely on noisy model inference for joint
information, our procedural paradigm provides full control over the semantics
and geometry of each part and joint, thereby ensuring the accuracy of physical
properties.
By configuring the pipeline with specific parameters, our programming rules can
generate highly sophisticated articulation trees.
With no upper limit on tree size, our pipeline is capable of constructing
objects that extend beyond the scope of existing datasets or reconstruction
from real scenes.
Beyond these, our hybrid asset pipeline seamlessly integrates procedurally
generated meshes with curated dataset assets during the assembly of articulated
objects, ensuring both the quality and diversity of the final mesh.

To demonstrate the superiority of our pipeline, we conduct evaluations in both
physical property and mesh quality, comparing our results with human-annotated
datasets and state-of-the-art generative methods.
Physical properties can only be evaluated in simulation, so we use
Sapien\cite{Xiang_2020_SAPIEN} to record the movement of each joint and invite
human annotators to rate the fidelity of videos.
Regarding mesh quality, we rendered both normal maps and RGB images for each
articulated object, constructing paired samples for comparison.
We leveraged vision language models (VLMs) as evaluators\cite{Wu_2024_CVPR} to
facilitate large-scale testing.
Results show that our pipeline provides results slightly better than datasets
and defeats state-of-the-art methods in all aspects.
Furthermore, we utilized our synthetic data to train generative models, thereby
facilitating subsequent scaling efforts.

In summary, our contributions are three-folded:
\vspace{-0.5em}
\begin{itemize}
    \vspace{-0.25em}
    \item We propose Infinite Mobility, a novel pipeline for synthesizing
          high-fidelity articulated objects through procedural generation boosted by 3D
          datasets annotated with part information.
          \vspace{-0.3em}
    \item We demonstrate that our method can produce results that excel current
          state-of-the-art methods and are comparable to human-annotated datasets in both
          physics property and mesh quality.
          \vspace{-0.3em}
    \item We show that our synthetic data can be used as training data of
          existing generative models, enabling next-step scaling.
\end{itemize}

\section{Related Work}
\label{sec:related_work}
\begin{figure*}[h]
    \vspace{-2em}
    \centering
    \includegraphics[width=1\textwidth]{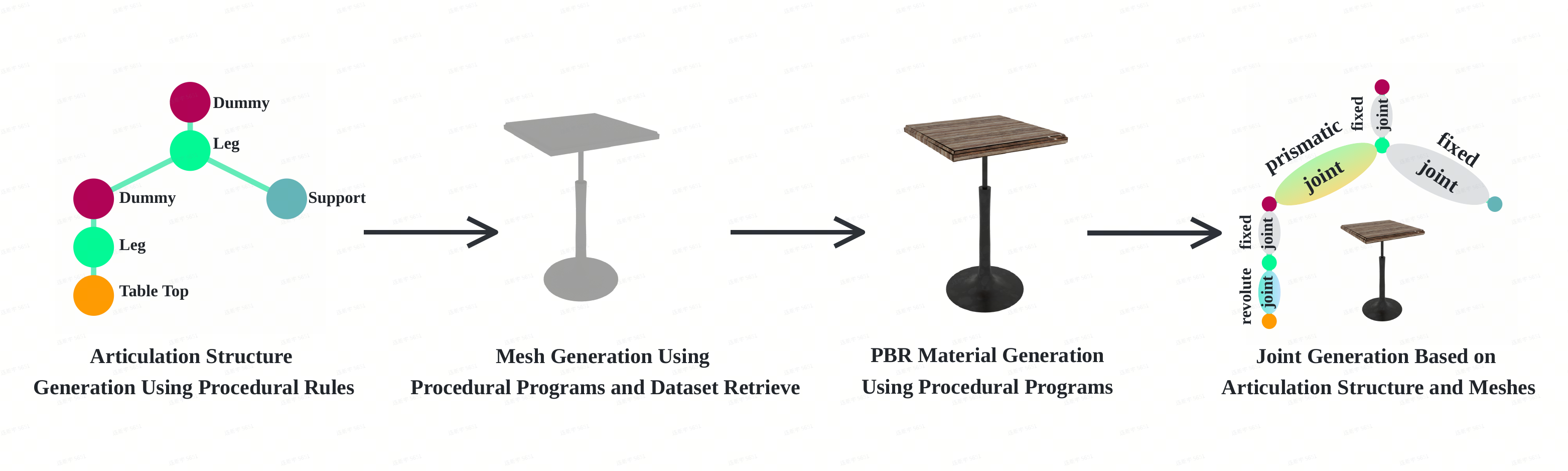}
    \vspace{-3em}
    \caption{
    The whole pipeline can be devided into four parts: articulation tree structure generation, geometry generation, material generation and joint generation. 
    }
    \label{fig:workflow}
\end{figure*}
Both procedural generation and articulated objetcs have been in public view for decades.
In this section, we compare the unique focuse of our work with existing procedural generators and reveal the vulnerability of other articulated objects data sources.
\paragraph{Procedural Generation For Machine Learning} 
Procedural generation has been broadly used in providing training data and building evaluation environments.
Since its debut in the 1980s, procedural generation quickly became a popular trick for game developing.
Agents\cite{MineStudio} playing such games as \textit{Rogue}, \textit{Minecraft}, \textit{Diablo III} and \textit{Civilisation VI} have attracted much attention. 
Procgen Benchmark\cite{pmlr-v119-cobbe20a} provides 16 procedrally generated game environments for multiple tasks.
In reinforcement learning, domain randomization\cite{8202133,weng2019DR} is a simple form of procedural generation and one way to mitigate overfitting in machine learning.
To prepare more complex scenes for embodied AI, ProcTHOR \cite{ProcTHOR} built a procedural system with more than 1000 static objects.
Pushing the boundary of procedural generation, Infinigen\cite{infinigen2023infinite,infinigen2024indoors} use purely programming rules with layout constraint solvers and achieve remarkable results.
All those works are mostly focused on generating scenes with static objects but ignoring more fine-grained interactive details of objetcs.  

\paragraph{Articulated Object Datasets} 
Articulated object datasets are a category of datasets that contain deformed mesh parts annotated with fine-grained hierarchical articulation information.
Various datasets have been proposed to provide assets in similar formats\cite{chang2015shapenet,Mo_2019_CVPR, RBO,10.1145/3130800.3130811,li2024rocaproboticdatacollection,Xiang_2020_SAPIEN,liu2022akb48realworldarticulatedobject,kim2024parahome,morlans2024aograsparticulatedobjectgrasp,NEURIPS2022_3b3a83a5}. 
ShapeNet\cite{chang2015shapenet} and PartNet\cite{Mo_2019_CVPR} are two datasets that pioneered to provide relatively large-scale part annotations either labelled in a combined pipeline or manually and PartNet\cite{Mo_2019_CVPR} has begun to offer hierarchically semantic information.
Based on these two datasets, PartNet-Mobility\cite{Xiang_2020_SAPIEN} selected 46 categories of articulated objects and endowed them with high-quality articulation information.
Extended from PartNet-Mobility, AO-Grasp\cite{morlans2024aograsparticulatedobjectgrasp} focused on grasping interaction with articulated objects and labelled a subset of PartNet-Mobility\cite{Xiang_2020_SAPIEN} with related data.
Those datasets, through some introduced annotation algorithms, still require massive labor for annotation and are limited in scale.
Instead of being derived from existing asset datasets, MultiScan\cite{NEURIPS2022_3b3a83a5}, AKB48\cite{liu2022akb48realworldarticulatedobject}, ParaHome\cite{kim2024parahome}, RoCap\cite{li2024rocaproboticdatacollection} and RBO\cite{RBO} consider from a real-to-sim perspective.
The first three use real-world scans or images to reconstruct meshes with articulation information, while the latter document interactions with interactable objects for articulation.
However, those digital twins that were reconstructed are not necessarily equipped with satisfactory mesh topology and articulation information.
Apart from this, the real-to-sim paradigm can hardly supply objects with sophisticated inner articulation structures or have not been made in reality.
\paragraph{Articulated Object Generation.}
\vspace{-0.2in}
Some approaches to creating articulated objects are quite similar with those used to create datasets.
Part of real-to-sim methods\cite{chen2024urdformer,jiayi2023paris,mandi2025realcode,le2025articulateanything,qiu2025articulate_anymesh} reconstruct whole object mesh and deploy segmentation algorithms to obtain part with single functionality and rely on articulation perception algotithms for articulation.
Others\cite{10.1109/ICRA46639.2022.9812272,weng2024neural,deng2024articulate,uzolas2024template} reconstruct object as implicit representation and retrieve desired parts after articulation perception.
These approaches need accurate articulation perception models for high-fidelity results.
Earlier ones\cite{chen2024urdformer,jiayi2023paris, 10.1109/ICRA46639.2022.9812272,weng2024neural,deng2024articulate,uzolas2024template} rely on their on models which depend on data-driven traning and the newest ones\cite{mandi2025realcode,qiu2025articulate_anymesh,le2025articulateanything} resort to VLMs or LLMs which are not specialized in this task for results.
Attempts have also been made to generate articulated objects from real-world simulation and virtual simulators.
Those routes\cite{jiang2022ditto,song2024reacto,zhong20233dimplicittransportertemporally,screwnet} usually use videos or object in different poses to infer articulation information.
However, the heavy labour and the noise in the real-world or virtual simulation data mired those ideas.
Generative\cite{lei2023nap,liu2024cage} models have also been used in this area.
Built upon ScrewNet\cite{screwnet}, NAP\cite{lei2023nap} design a new articulation tree parameterization and then apply a DDPM\cite{ho2020denoisingdiffusionprobabilisticmodels} as generation backbone. 
CAGE\cite{liu2024cage} continues to deploy DDPM\cite{ho2020denoisingdiffusionprobabilisticmodels} but with a more novel graph attention as network model and inject control signals for better usages.
Nonetheless, limited training data still hinders generation with high standards.

\section{Methodology}
\label{sec:method}
\subsection{Preliminary}
\begin{figure}[h]
    \vspace{-1em}
    \centering
    \includegraphics[width=1\textwidth]{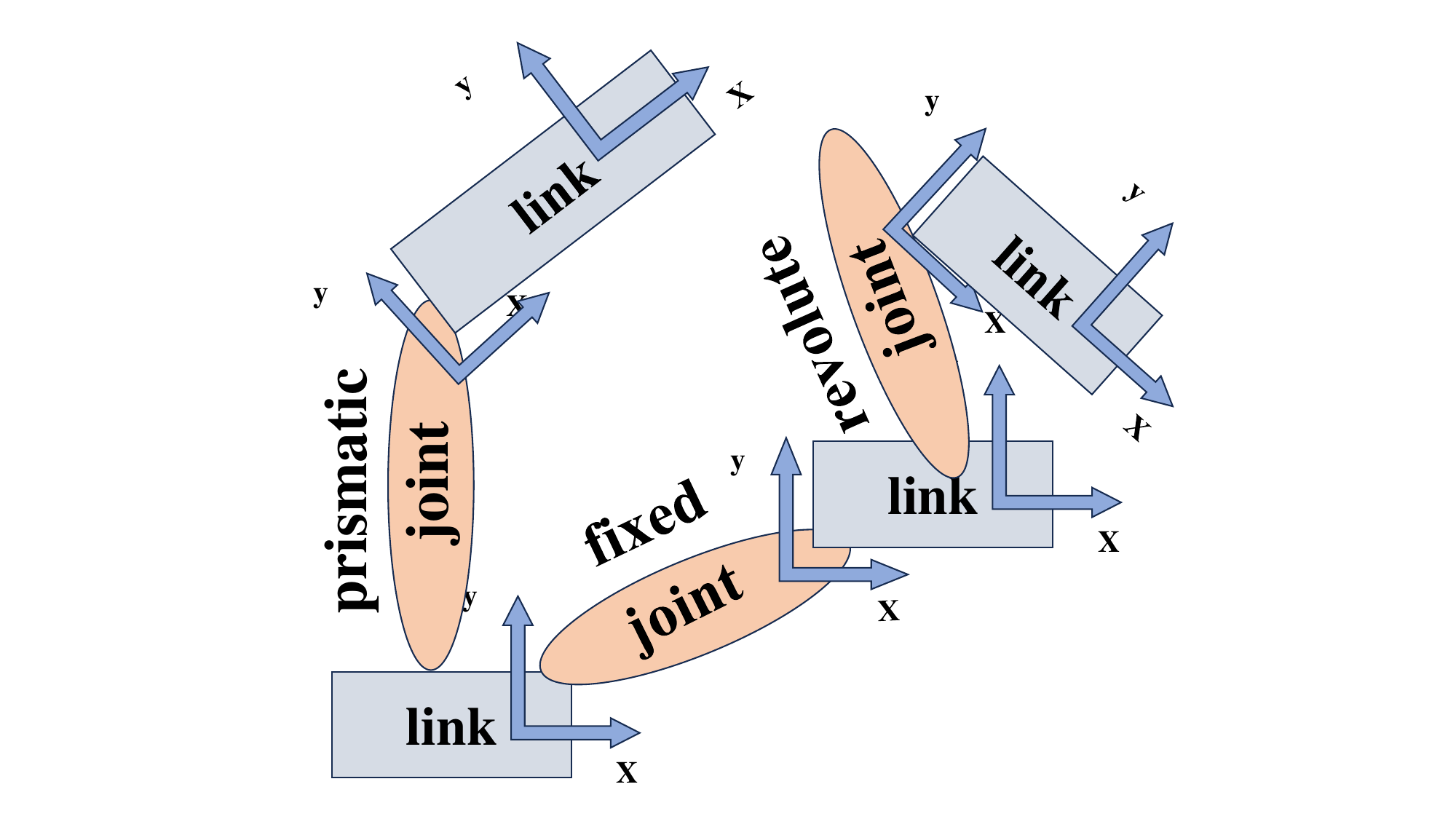}
    \vspace{-1em}
    \caption{
    Structure of the URDF file. 
    Each link is a part of the object, which is represented as a textured mesh in our case.
    Each joint connects two links and describes the articulation structure between them.
    }
    \label{fig:urdf_example}
    \vspace{-1.9em}
\end{figure}

\begin{figure}[h]
    \vspace{-1.2em}
    \centering
    \includegraphics[width=1\textwidth]{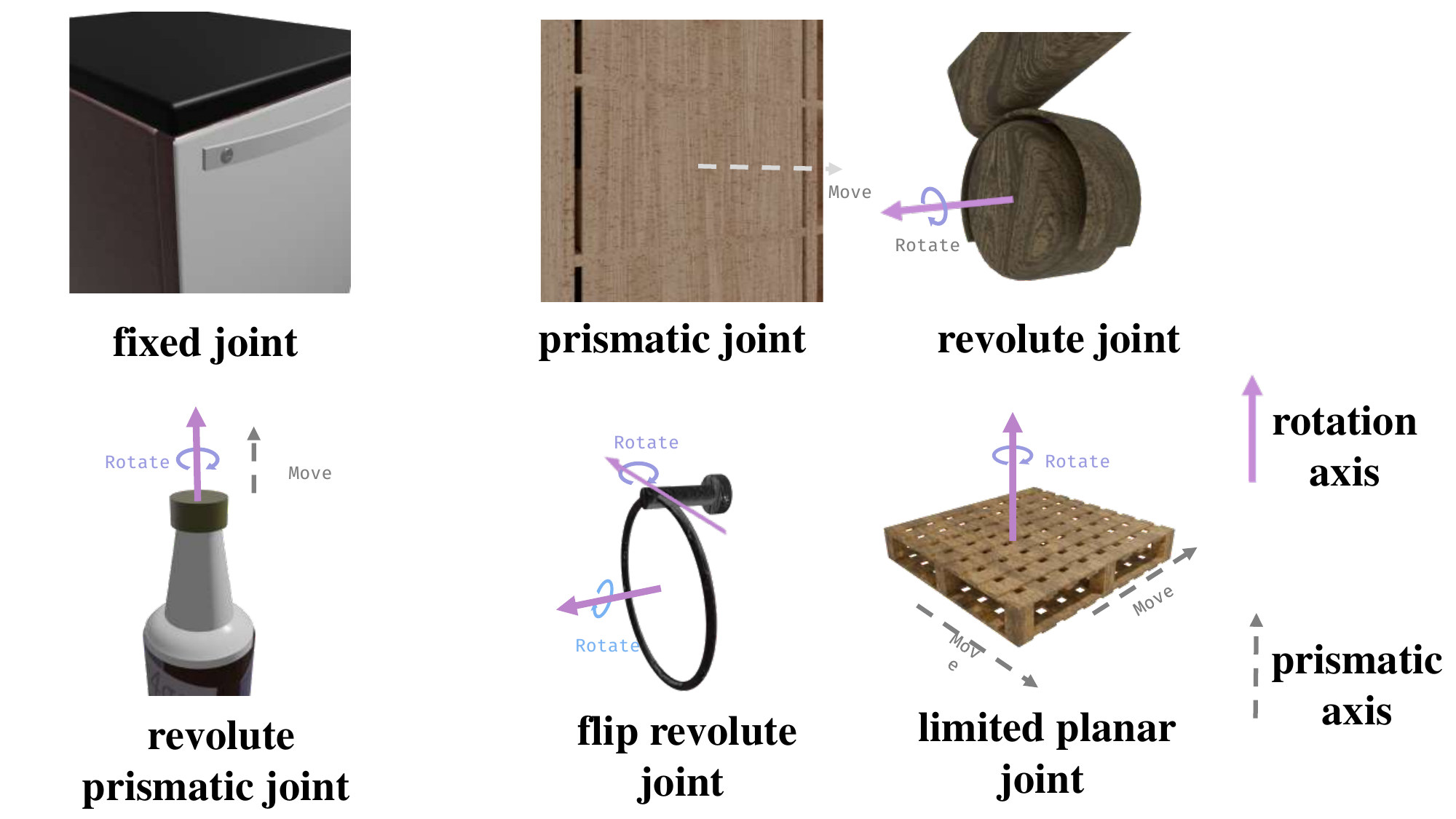}
    \caption{
    We implement $6$ kinds of joints in our articulated objects.
    First three are simple joints, and the last three are compound joints.
    }
    \label{fig:joints}
    \vspace{-0.5em}
\end{figure}

Articulated objects constitute a class of objects that consist of multiple rigid bodies connected by various joints.
We refer to URDF (Unified Robot Description Format) as the standard format for describing articulated objects.
Each rigid body is described as a link, and each joint is a connection between two links.
With such structure, an articulated object can be described as a tree structure, where each node represents a link and each edge represents a joint.
A simple structure described by URDF is shown in Figure~\ref{fig:urdf_example}.
In URDF, each link is described by its shape which can be either one of its predifined shapes or path to meshes stored in common formats.
Joints in URDF are described by their origin, type, axis, limits, and dynamics properties.
With regularly supported joint types like revolute, prismatic, continuous and fixed URDF is capable to portray simple movements with low DOFs.
To describe more complex movements, URDF allows for the definition of compound joints, which are joints that are composed of multiple simple joints and dummy links with no shape specified.
Such compound joints with simple joints connected one by one enable the description of complex movements with high DOFs.
Joints we implemented are illustrated in Figure~\ref{fig:joints}. 
Little work has been done to generate articulated objects with compound joints which are common in daily objects.

\subsection{Articulated Object Generation}
\begin{figure*}[h]
    \centering
    \includegraphics[width=1\textwidth]{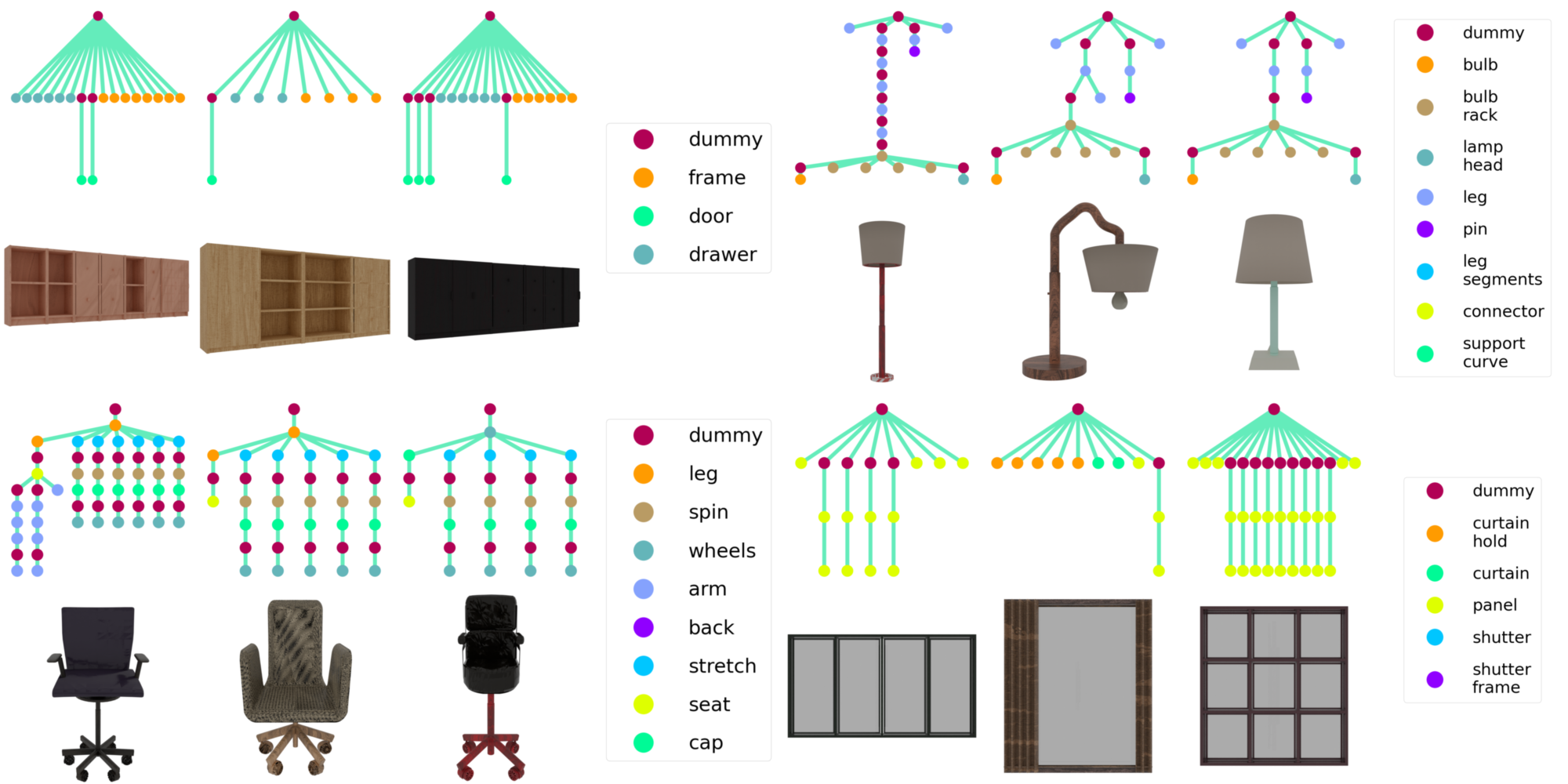}
    \caption{Examples of our generated cabinets, chairs, lamps and windows.
    For each object, we display the textured mesh with the corresponding articulation tree above.
    The generated objects are diverse in both shapes and articulation structures.
}
    \label{fig:joint_generation}
\end{figure*}

For the construction of articulated objects, both geometric and articulation data must be synthesized.
Contrary to real-to-sim methods\cite{RBO, jiang2022ditto,jiayi2023paris,liu2022akb48realworldarticulatedobject,NEURIPS2022_3b3a83a5} and approaches using given 3D asset datasets\cite{Xiang_2020_SAPIEN,wang2019shape2motion,10.1145/3355089.3356573} which extract articulation information out of an given object, our pipeline is much like those generative models for articulated objects\cite{lei2023nap,liu2024cage} which build the articulation structure first and then fill each node with meshes.
Our work generates all articulation structures through a collection of purposefully designed programs that utilize randomly sampled input parameters. The codebase employed for result generation is systematically organized into category-specific generators, thereby enhancing usability and facilitating more efficient implementation.
With most shape related parameters controllable by users and joint related parameters sampled from range calculated out of shape parameters, we can ensure both highly flexible mesh control and high-fidelity articulation structure generation to be achieved by our pipeline. 
Workflow of our pipeline is summarized in Figure~\ref{fig:workflow}.
\vspace{-0.5em}
\paragraph{Articulation Tree Structure Generation}
In our pipeline, articulation tree structures are formulated as adjacency lists.
With each node connected to only one parent joint, we can easily constraint a tree structure to be formed.
For each daily-life category, there exists a general understanding of proper part attachments.
Leveraging this understanding, we can establish a category-specific growth strategy for articulation trees.
For every node with certain semantics, we define its requested children and plausible branches that can be attached to it.
Those requested children depict the basic structure of the object and are necessary for the object to be complete like the legs and table top of a table.
Plausible branches are optional and can be added to the object to increase its diversity like buttons of microwave.
Commencing from a root node, which signifies the object's origin, nodes can be recursively appended to the tree structure.
For nodes amenable to attachments, the addition of new branches is dictated by discrete random parameters.
Through recursion, a plausible tree structure is derived for each sample.
\vspace{-1.2em}
\paragraph{Geometry Generation} 
Our geometry generation process is implemented via two approaches: procedural mesh generation and mesh retrieval with procedural refinement.
For both methods, we first determine the bounding box of each node based on its semantics and generated meshes of other nodes.
We will introduce our mesh retrieve approach in detail later in section~\ref{subsec:mesh_retrieve}.
For procedural mesh generation, our pipeline is based on Infinigen\cite{infinigen2023infinite,infinigen2024indoors} augmentated with our elabrate mesh creation programs.
\vspace{-1.2em}
\paragraph{Material Generation}
Materials used in our work are based on Blender shader node system.
The material generation process is based on Infinigen\cite{infinigen2023infinite,infinigen2024indoors} boosted by collected crafted materials.
To add more randomness, we modify diffuse, specular, roughness and normal maps of materials with random noise. 
For exportation, we resort to API created by Infinigen to save materials and textures in a format that can be easily imported into simulation tools.
\vspace{-1.2em}
\paragraph{Joint Generation}
Once the articulation tree structure is generated, the joint type for each edge is determined based on its parent and child nodes.
The physical properties of each joint, including its axis, position, and motion range, are derived from its type and the overall generated meshes.
This approach enables the generation of articulation trees that are both topologically diverse and physically high-fidelity. 
Result samples are shown in Figure~\ref{fig:joint_generation}.
\subsection{Procedural Mesh Retrieve}
\label{subsec:mesh_retrieve}
\begin{figure}[h]
    \centering
    \includegraphics[width=1\textwidth]{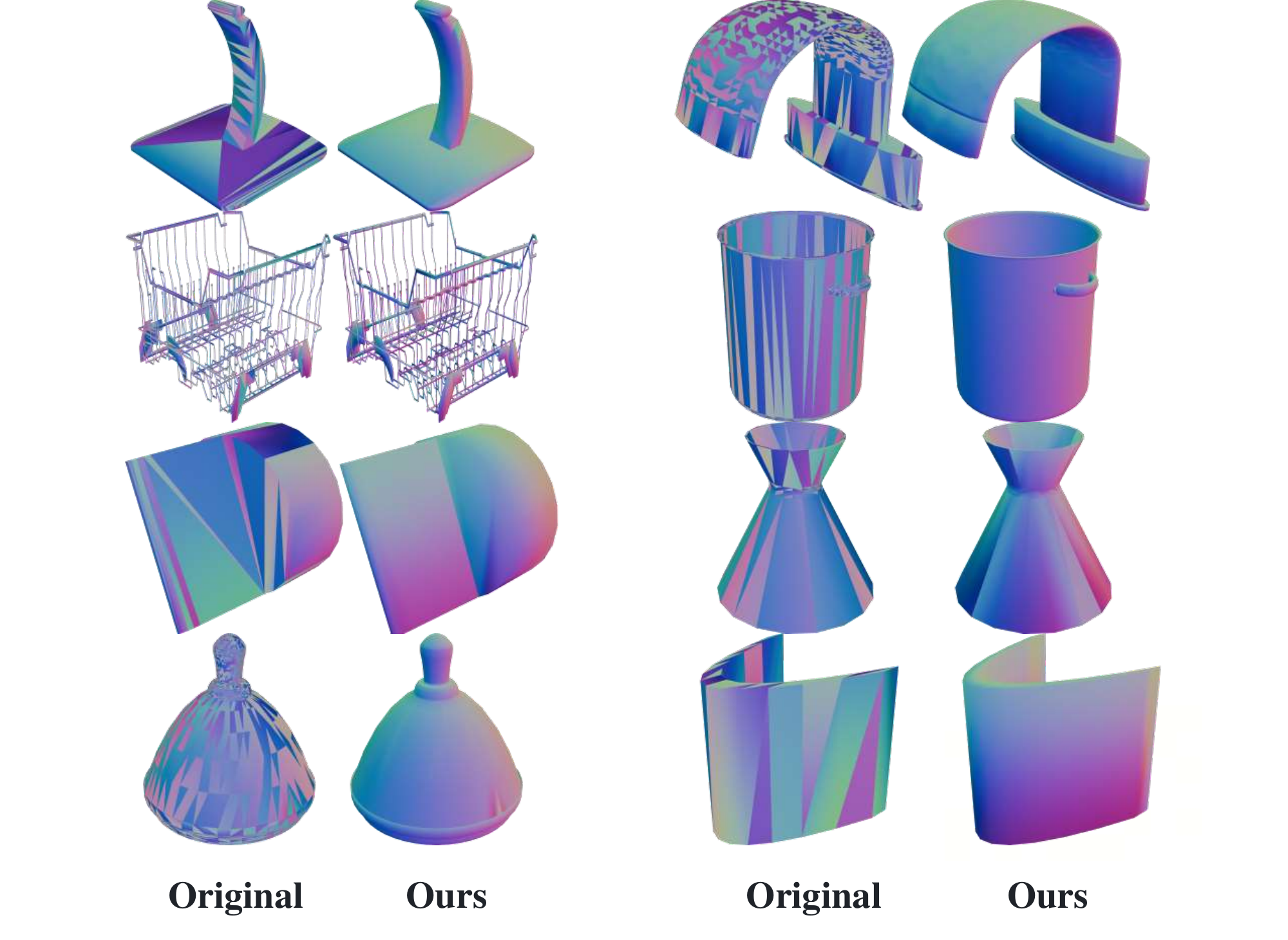}
    \caption{
    Original parts from PartNet and ShapeNet bear many back faces and thus their normals are highly irregular.
    We flip these back faces in Blender using recalculate normal function followed by human repair and ensure the meshes bear consistent outward-facing normals.
    }
    \label{fig:dataset_comparison}
    \vspace{-1em}
\end{figure}

\begin{figure}[h]
    \centering
    \includegraphics[width=1\textwidth]{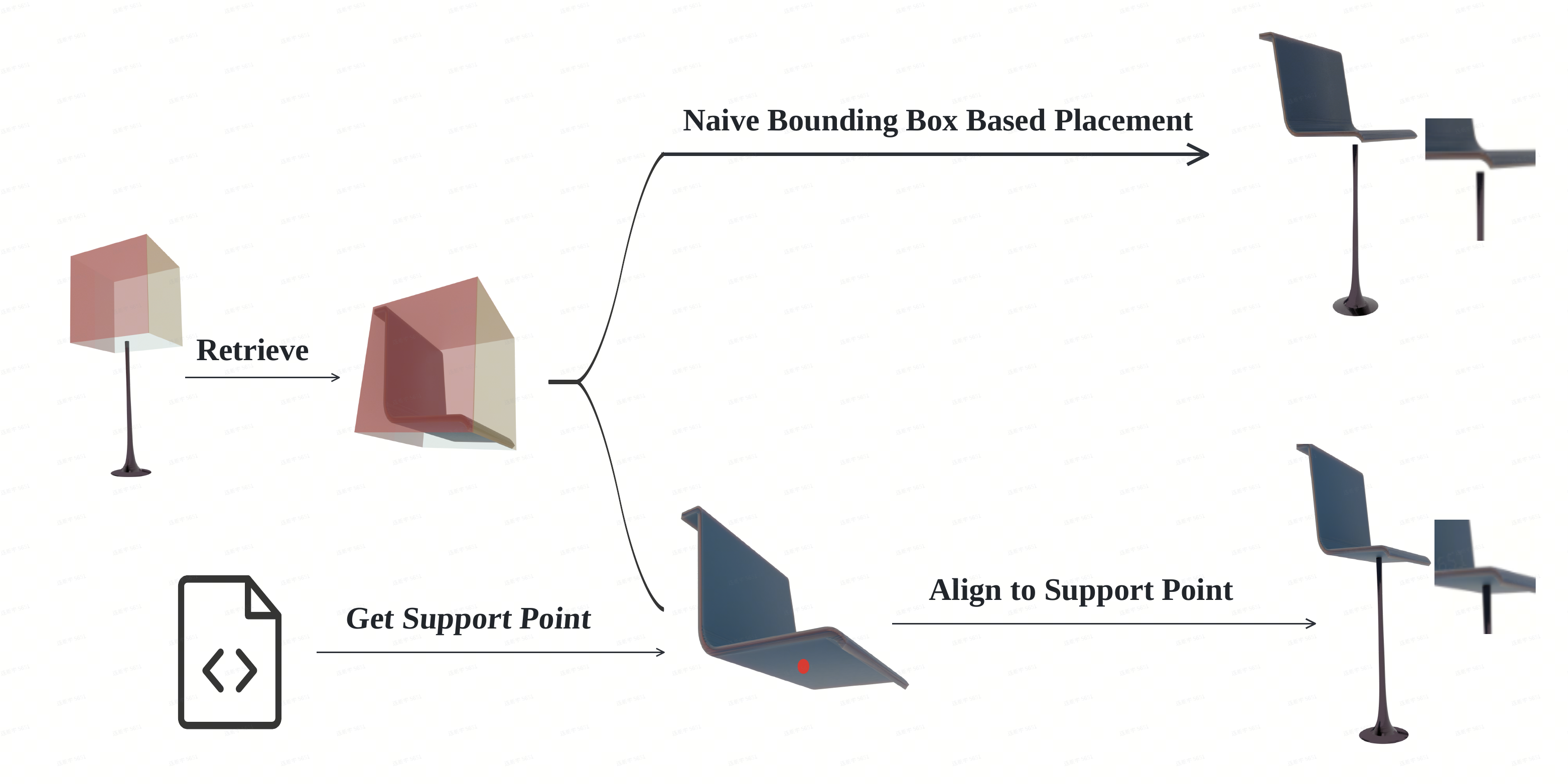}
    \vspace{-1.6em}
    \caption{
    We adopt support point-based placement to position the retrieved part on the object.
    Naive bounding box-based placement may create a gap between the retrieved part and the object. Our approach guarantees a seamless connection.
    }
    \label{substitution_process}
\end{figure}

As previously, meshes used in our pipeline can be retrieved from our curated dataset.
This section outlines the methodology behind our dataset construction and details the mesh retrieval process.
\paragraph{Dataset Construction}
3D asset daasets like ShapeNet\cite{chang2015shapenet} and PartNet\cite{Mo_2019_CVPR} use segmentation algorithms to obtain part meshes and label them with various annotations.
However, those meshes in such datasets are not consistent in quality and part with single functionaliy is often segmentated into many pieces, resulting in bad visual fidelity and obstacles when applying new materials.
We concentrate on 22 part labels which are most commonly interacted when agents are trained to perform tasks and quary those parts according to labels.
By inspecting and fixing those meshes by huamn labor, we obtain a curated dataset containing hundreds of meshes with requisite information.
Comparisons between meshes from our dataset and meshes in the original dataset are shown in Figure~\ref{fig:dataset_comparison}.
\begin{figure*}[h]
    \centering
    \includegraphics[width=1\textwidth]{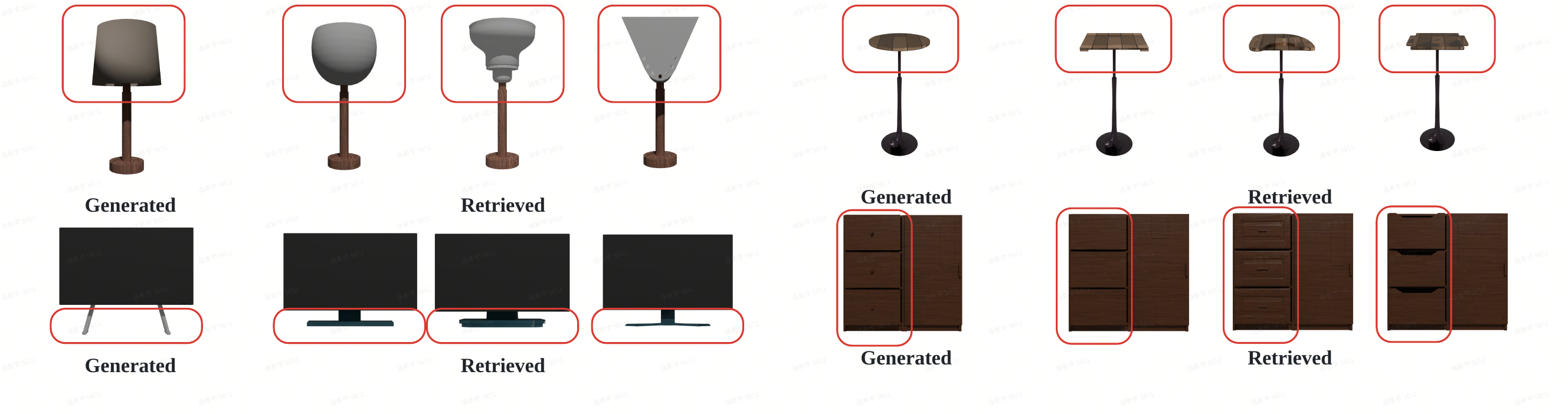}
    \vspace{-3em}
    \caption{
    We use two methods to obtain the geometry of each part of an object. 1) Blender python API created meshes. 2) Parts retrieved from our carefully curated and processed dataset.
    For each group of objects, the left one is generated by Blender python API, while the right $3$ objects are obtained by retrieving parts from curated PartNet-Moblity.
    Parts obtained from both methods can be seamlessly joined, and it improves the diversity of generated shapes by adopting both methods to obtain part geometry.
    }
    \label{fig:mesh_substitution}
\end{figure*}
\vspace{-1.2em}
\paragraph{Mesh Retrieve with Procedural Refinement}
According to the articulation tree generated, we can determine the semantics of every node and thus retrieve mesh from dataset mentioned above.
For most existing methods, retrieved meshes are only transformed to fit bounding boxes.
However, if new meshes are placed directly into required bounding boxes, there would be parts with complex shapes that either suspend in the air or penetrate into other parts.
To avoid such situations, we only use bounding box for initial placements and then define each part label with procedurally defined critical supporting points.
By aligning meshes with those points, we can ensure that the final mesh is physically plausible and visually appealing.
Example of such situation and our fix is shown in Figure~\ref{substitution_process}.
Qualitative results are shown in Figure~\ref{fig:mesh_substitution}.

\subsection{Ensure Physical Plausibility}
\begin{figure*}[h]
    \centering
    \includegraphics[width=1\textwidth]{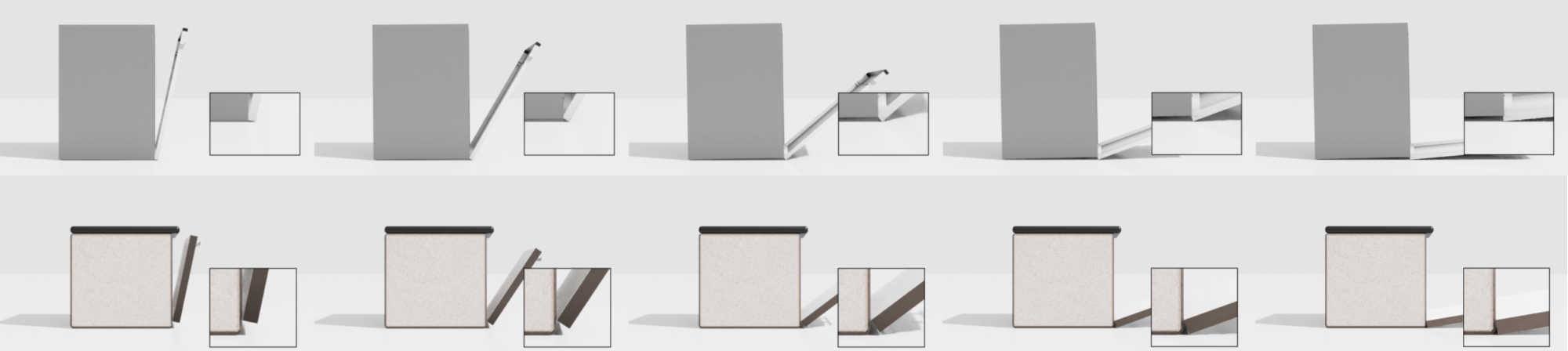}
    \vspace{-2em}
    \caption{
    We carefully position each part and optimally configure the joint parameters to guarantee that no collisions occur either among the parts themselves or between the object and the ground during its articulation.
    The top row is a dishwasher from PartNet-Mobility.
    When the dishwasher door is opened, its base collides with the ground, forcing the main body to tilt upwards.
    While the dishwasher generated by our method in the bottom row does not have this problem.
    }
    \label{fig:physical_plausibility}
\end{figure*}

Creating fancinating meshes combined in a tree structure with proper semantics is able to ensure the basic fidelity of the generated objects.
However, to adopt those objects in simulation tools, a series of amendments need to be adopted to conquer following problems we observed. 
\vspace{-1.2em}
\paragraph{Problem: Ill-designed parts lead to unreasonable collisions with rigid ground.} Some artists make models differnet from reasonable real world objects for better appearance and simplicity.
This phenomenon happens frequently when it comes to the base part of an object.
Objects created this way suffer from problems when parts in lower positions are articulated with joints moving downward.    
    \\\textbf{Modification:} For those categories, we either build a base to position the articulated part further from the ground, or we modify the articulated parts by shrinking their sizes and limiting their motion ranges to a proper number.
    \vspace{-1.2em}
    \paragraph{Problem: Insufficient gap between parts cause unstable movements of articulated joints.}
    It's common to build meshes as a whole with no gaps between parts for convenience when modeling.
    However, this makes it difficult for simulation to handle the collisions when such meshes are directly divided into parts and imported.
    When meshes are imported into simulation, different simulators have differnet strategies for collision detection.
    But those methods hardly use original meshes as the real collidors but resort to other approximations for efficiency.  
    Such practice usually brings wrong collisions into the simulation of articulated joints even if those meshes are not overlapping. 
    \\\textbf{Modification:} For any parts articulated by joints, we additionally bring in gaps of 2\% of original scales of meshes.
    \vspace{-2em}
    \paragraph{}We present simulation results of one example and our amendment result in Figure~\ref{fig:physical_plausibility}.
It's worth mentioning that problems mentioned above are quite frequent in articulated objects from other sources and appear in commonly used datasets like PartNet-Mobility\cite{Xiang_2020_SAPIEN}.

\section{Experiment}
\label{sec:experiment}

Our pipeline constitutes a high-quality data source similar to existing datasets like PartNet-Mobility\cite{Xiang_2020_SAPIEN} and surpass generative model like NAP\cite{lei2023nap} and CAGE\cite{liu2024cage}.
In this section, we evaluate the quality of our generated results, train a generative model on our results and verify the application of our results in embodied AI.
\subsection{Evaluations of Articulated Objects}
\begin{figure}[!h]
    \centering
    \includegraphics[width=1\textwidth]{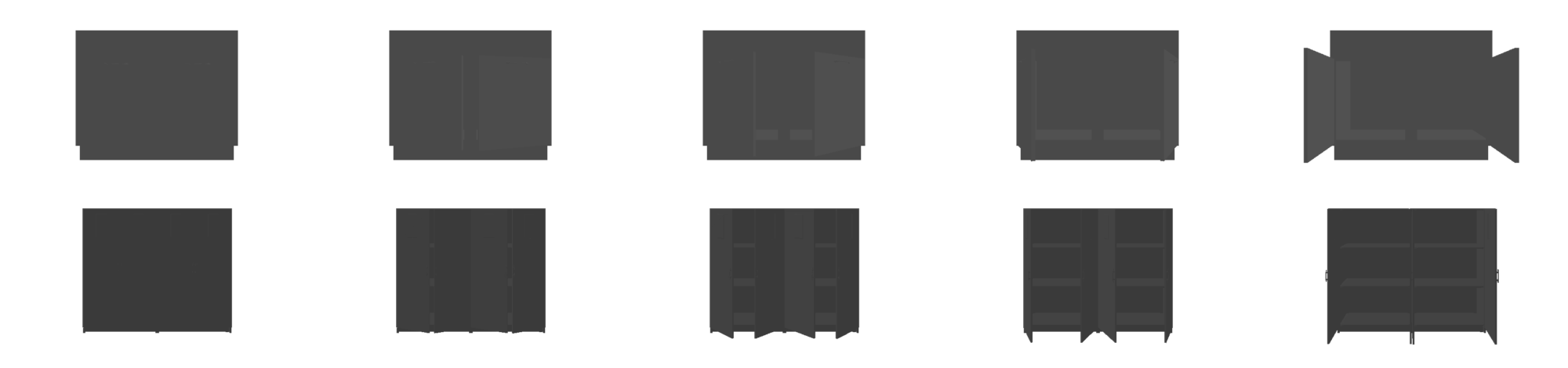}
    \caption{
    Human evaluators observe textureless videos of our generated articulated objects (lower row) and those from PartNet-Mobility (upper row), subsequently determining which exhibits superior motion structure.
    }
    \label{fig:joint_evaluation}
\end{figure}

We first assess the quality of our articulated objects by comparing them against the PartNet-Mobility dataset\cite{Xiang_2020_SAPIEN} and samples produced by state-of-the-art generative models \cite{lei2023nap,liu2024cage}. 
To evaluate joint fidelity, we conduct human evaluations in which 10 participants review 50 randomly selected object pairs from each category, judging joint performance based on movement videos.
One sample pair is displayed in Figure \ref{fig:joint_evaluation}.
In addition, visual language models (VLMs) are employed to assess mesh quality. 
Following a recent evaluation paradigm \cite{Wu_2024_CVPR}, we utilize GPT-4v as an evaluator by feeding it normal maps and RGB images, thereby enabling large-scale, unbiased testing that rates both the geometry and texture of the meshes.
We present the results in Table \ref{tab:ours_vs_partnet} and Table \ref{tab:ours_vs_cage}.

Complexity and diversity of our results and PartNet-Mobility\cite{Xiang_2020_SAPIEN} are also measured through quantitative metrics.
For complexity, we calculated average joint numbers of each category from both data sources.
For diversity, we calculated variances of joint number and average Tree-Edit-Distance between each pair of articulation trees in a category.
Results are shown in Table \ref{tab:metrics}.

We further compare the generation time of our pipeline with NAP\cite{lei2023nap} and CAGE\cite{liu2024cage}.
Results are shown in Table \ref{tab:Speed}.

\begin{figure}[!h]
    \centering
    \includegraphics[width=1\textwidth]{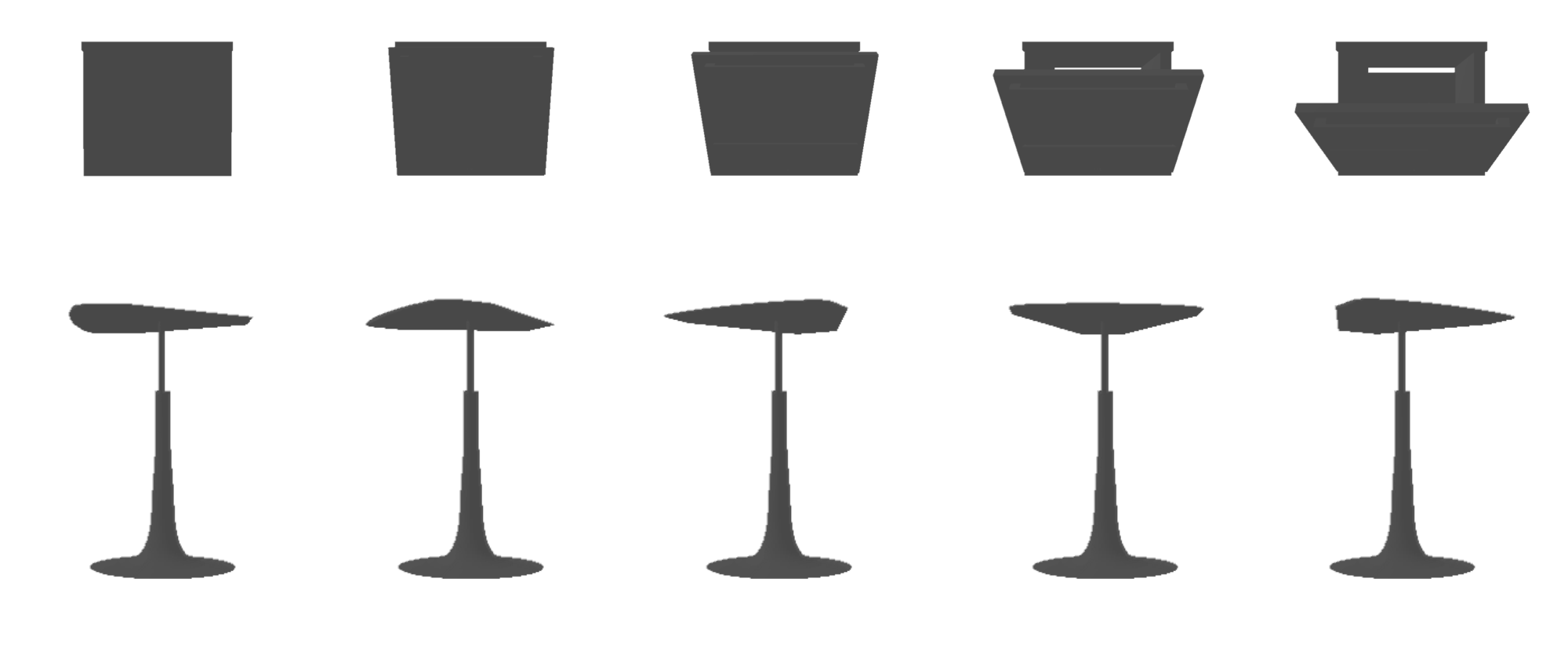}
    \caption{Examples of articulated objects generated by CAGE trained on our results.
    The generated object has accurate geometry and reasonable motion structure.}
    \label{fig:our_cage}
\end{figure}




\begin{table*}[t]
    \centering
    \caption{
    Compare articulated objects generated by our method and those from PartNet-Mobility in terms of  kinematic articulation, geometric accuracy, and textural fidelity.
    We report the comparative success rates of our method versus PartNet-Mobility.
    ``Equal'' denotes that the human evaluator or GPT assess the quality of the two objects as equivalent.
    }
    \label{tab:ours_vs_partnet}
    \resizebox{1\textwidth}{!}{ 
    \begin{tabular}{lcccccccccc}
        \toprule
        \multirow{2}{*}{Category} & \multicolumn{3}{c}{Human Evaluation (Articulcation)} & \multicolumn{3}{c}{GPT Evaluation (Geometry)} & \multicolumn{3}{c}{GPT Evaluation (Texture)} \\
        \cmidrule(lr){2-4} \cmidrule(lr){5-7} \cmidrule(lr){8-10}
        & Ours (\%) & PartNet-Mobility (\%) & Equal (\%) & Ours (\%) & PartNet-Mobility (\%) & Equal (\%) & Ours (\%) & PartNet-Mobility (\%) & Equal (\%) \\
        \midrule
        All Categories & \textbf{8.76} & 4.63 & 86.61 & \textbf{64.18} & 35.45 & 0.37 & \textbf{84.81} & 14.44 & 0.74 \\
        Bottle & \textbf{9.64} & 4.42 & 85.94 & \textbf{78.00} & 21.56 & 0.44 & \textbf{91.56} & 8.44 & 0.00 \\
        Dishwasher & \textbf{15.17} & 4.99 & 79.84 & \textbf{71.43} & 28.57 & 0.00 & \textbf{85.45} & 14.55 & 0.00 \\
        Display TV & \textbf{8.20} & 5.40 & 86.40 & 35.14 & \textbf{64.86} & 0.00 & \textbf{58.86} & 40.54 & 0.60 \\
        Door & \textbf{7.20} & 3.20 & 89.60 & \textbf{74.60} & 25.08 & 0.32 & \textbf{79.37} & 20.32 & 0.32 \\
        Fridge & \textbf{3.00} & 0.60 & 96.40 & \textbf{69.79} & 29.95 & 0.26 & \textbf{87.34} & 12.66 & 0.00 \\
        Lamp & 7.40 & \textbf{8.00} & 84.60 & \textbf{75.13} & 24.11 & 0.76 & \textbf{77.48} & 21.62 & 0.90 \\
        Microwave & \textbf{7.39} & 2.20 & 90.42 & \textbf{62.39} & 37.61 & 0.00 & \textbf{64.10} & 35.90 & 0.00 \\
        Oven & \textbf{8.76} & 1.39 & 89.84 & 26.34 & \textbf{73.66} & 0.00 & \textbf{56.84} & 42.74 & 0.43 \\
        Pot & \textbf{4.20} & 0.40 & 95.40 & \textbf{54.67} & 45.33 & 0.00 & \textbf{82.22} & 17.33 & 0.44 \\
        Table & \textbf{13.57} & 4.79 & 81.64 & \textbf{72.89} & 26.89 & 0.22 & 24.67 & \textbf{75.11} & 0.22 \\
        Tap & \textbf{9.60} & 0.20 & 90.20 & \textbf{90.00} & 10.00 & 0.00 & \textbf{93.11} & 6.89 & 0.00 \\
        Toilet & \textbf{22.86} & 0.00 & 77.14 & \textbf{59.33} & 40.67 & 0.00 & \textbf{87.56} & 12.44 & 0.00 \\
        Cabinet & \textbf{9.40} & 2.60 & 88.00 & 47.33 & \textbf{52.67} & 0.00 & \textbf{93.65} & 6.35 & 0.00 \\
        Window & 3.21 & \textbf{6.41} & 90.38 & \textbf{76.81} & 23.19 & 0.00 & \textbf{75.93} & 24.07 & 0.00 \\
        Chair & 1.99 & \textbf{4.57} & 93.44 & \textbf{66.29} & 32.20 & 1.52 & \textbf{78.52} & 21.48 & 0.00 \\
        \bottomrule
    \end{tabular}}
\end{table*}
\floatsetup[table]{capposition=top}
\newfloatcommand{capbtabbox}{table}[][\FBwidth]

\begin{table*}[htbp]
    \vspace{-10em}
    \resizebox{1\textwidth}{!}{
    \begin{floatrow}
    \capbtabbox{
    \resizebox{0.6\textwidth}{!}{ 
    \begin{tabular}{lcccccc}
        \toprule
        \multirow{2}{*}{Category} & \multicolumn{3}{c}{Human Evaluation (Articulation)} & \multicolumn{3}{c}{GPT Evaluation (Geometry)} \\
        \cmidrule(lr){2-4} \cmidrule(lr){5-7}
        & Ours (\%) & CAGE (\%) & Equal (\%) & Ours (\%) & CAGE (\%) & Equal (\%) \\
        \midrule
        All Categories & \textbf{22.95} & 5.60 & 71.45 & \textbf{79.26} & 20.74 & 0.00 \\
        Table & \textbf{19.04} & 7.62 & 73.35 & \textbf{87.50} & 8.33 & 4.17 \\
        Cabinet & \textbf{11.82} & 9.82 & 78.36 & 48.00 & \textbf{51.33} & 0.67 \\
        Dishwasher & \textbf{29.58} & 2.01 & 68.41 & \textbf{65.62} & 34.38 & 0.00 \\
        Microwave & \textbf{28.69} & 3.20 & 70.92 & \textbf{65.28} & 34.72 & 0.00 \\
        Oven & \textbf{22.24} & 2.20 & 77.56 & \textbf{69.79} & 29.95 & 0.26 \\
        Fridge & \textbf{29.28} & 1.99 & 68.73 & \textbf{84.31} & 15.69 & 0.00 \\
        \bottomrule
        \multirow{2}{*}{Category} & \multicolumn{3}{c}{Human Evaluation (Articulation)} & \multicolumn{3}{c}{GPT Evaluation (Geometry)} \\
        \cmidrule(lr){2-4} \cmidrule(lr){5-7}
        & Ours (\%) & NAP (\%) & Equal (\%) & Ours (\%) & NAP (\%) & Equal (\%) \\
        \midrule
        All Categories & \textbf{58.36} & 7.96 & 33.69 & \textbf{74.81} & 24.81 & 0.37 \\
        \bottomrule
    \end{tabular}
    }}{
    \caption{
    Compare articulated objects generated by our method, CAGE and NAP in terms of  kinematic articulation and geometric accuracy.
    }
    \label{tab:ours_vs_cage}
    }
    \begin{minipage}{1\textwidth}
    \vspace{10.5em}
    \capbtabbox{
    \resizebox{0.35\textwidth}{!}{ 
    \hspace{-0.3em}
    \begin{tabular}{lcc}
        \toprule
        Metrics & Ours  & PartNet-Mobility \\
        \midrule
        Average Joint Number & \textbf{12.32} & 5.91  \\
        Variance of Joint Number & \textbf{659.02} & 40.31   \\
        Tree Edit Distance & \textbf{78.62} & 3.88   \\
        \bottomrule
    \end{tabular}
    }}{
    \caption{
    Compare articulated objects generated by our method and those from PartNet-Mobility in terms of joint numbers and diversity of tree structures.
    }
    \label{tab:metrics}
    }
    \vspace{1em}
    \capbtabbox{
    \resizebox{0.35\textwidth}{!}{
    \begin{tabular}{lccc}
        \toprule
        Metrics & Ours & NAP & CAGE\\
        \midrule
        Time(s / object) & \textbf{0.46} & 2.04 & 1.96\\
        \bottomrule
    \end{tabular}
    }}{
    \caption{
    Compare generation time of our method, NAP and CAGE.
    }
    \label{tab:Speed}
    }
\end{minipage}
\end{floatrow}
    }
\end{table*}
    

\subsection{Training Generative Models}
To further scale the generation of articulated objects, we train generative models using results from our pipeline.
We use CAGE\cite{liu2024cage} as the generative model and train it with our generation results.
Different from the original CAGE trained using PartNet-Mobility\cite{Xiang_2020_SAPIEN} which is small in scale, we used 1000 samples per category for training.
We display two result samples in Figure \ref{fig:our_cage}.

\subsection{Embodied AI}
\begin{figure*}[!h]
        \centering
        \includegraphics[width=1\textwidth]{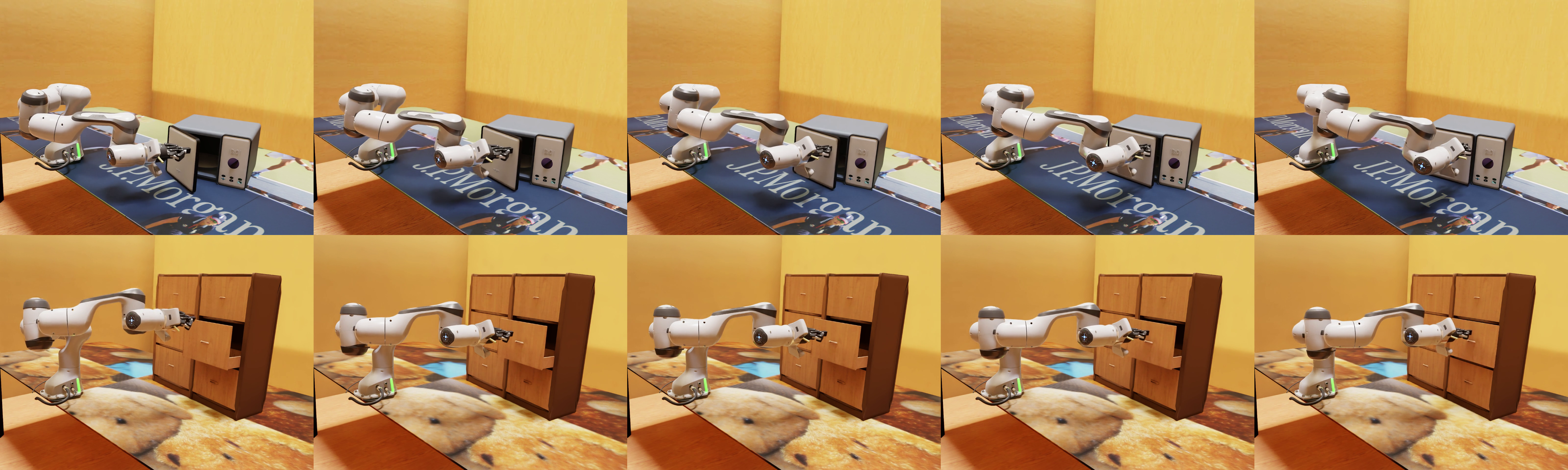}
        \caption{
        Kinematic sequences demonstrating a bimanual robotic manipulator interacting with our generated articulated objects within the Isaac Sim simulation environment.
        }
        \label{fig:isaac_sim}
    \end{figure*}

Our results can be imported in popular simulation environmets like Sapien\cite{Xiang_2020_SAPIEN}, Isaac Sim and Genesis\cite{Genesis}.
In isaac sim, we use our results to annotate waypoints for articulated objects and train agents using motion-planning algorithms.
We present samples in Figure \ref{fig:isaac_sim}.

\section{Limitations}
\label{sec:limitation}
At this stage, Infinite Mobility still has some limitations which could be explored in future works.
Since this version of Infinite Mobility is built on generators with largely human crafted rules, it require considerable efforts to be expended to other categories.
The automation of this process could probably be achieved by using LLMs with enough spatial intelligence and coding abilities.
Joints parameters like type, axis and motion range have been created with high quality now, but properties like friction, damping and motor strength are still missing.
Work could be done to infer those properties from meshes, materials and joint types.
\section{Conclusion}
\label{sec:conclusion}
In this paper, we present a procedural pipeline for generating high-fidelity articulated objects.
Quality of procedurally generated data from our work beat those from other models and rival best dataset now in both appearances and articulation properties.
Our work demonstrates the potential of procedural generation in creating not only appearance appealing but also structure diverse objects.

{\small
\bibliographystyle{ieee_fullname}
\bibliography{egbib}
}

\end{document}